\pdfoutput=1

\documentclass[11pt]{article}

\usepackage{ACL2023}

\usepackage{times}
\usepackage{latexsym}

\usepackage[T1]{fontenc}

\usepackage[utf8]{inputenc}

\usepackage{microtype}

\usepackage{inconsolata}

\usepackage{graphicx}
\usepackage{xcolor}
\usepackage{tikz}
\usepackage{hhline}

\title{Testing Causal Models of Word Meaning in GPT-3 and -4}

\author{Sam Musker\textsuperscript{1,2} \\
  \texttt{samuel\textunderscore musker@brown.edu} \\
   \\
  \textsuperscript{1}Department of Computer Science, Brown University \\
  \textsuperscript{2}Department of Philosophy, Brown University
  \\\And
  Ellie Pavlick\textsuperscript{1} \\
  \texttt{ellie\textunderscore pavlick@brown.edu} \\
  }

\begin{document}
\maketitle
\begin{abstract}

Large Language Models (LLMs) have driven extraordinary improvements in NLP. However, it is unclear how such models represent lexical concepts--i.e., the meanings of the words they use. This paper evaluates the lexical representations of GPT-3 and GPT-4 through the lens of HIPE theory, a theory of concept representations which focuses on representations of words describing artifacts (such as “mop”, “pencil”, and “whistle”). The theory posits a causal graph that relates the meanings of such words to the form, use, and history of the objects to which they refer. We test LLMs using the same stimuli originally used by \citet{Chaigneau:04} to evaluate the theory in humans, and consider a variety of prompt designs. Our experiments concern judgements about causal outcomes, object function, and object naming. We find no evidence that GPT-3 encodes the causal structure hypothesized by HIPE, but do find evidence that GPT-4 encodes such structure. The results contribute to a growing body of research characterizing the representational capacity of large language models.
\end{abstract}

\section{Introduction}

The success of large language models (LLMs) at generating human-like text has spurred a wave of recent work which aims to measure the extent to which such models have good representations of word meanings (i.e., lexical concepts). Such work has taken a variety of forms across multiple domains, but in general amounts to measuring the extent to which the conceptual associations encoded by LLMs match human associations. For example, prior work has shown that LLMs correctly associate physical objects with their properties and affordances \cite{Forbes:19,da2019cracking}; common nouns with the ontological categories \cite{da2019cracking,ettinger2020bert}, and entities with their salient characteristics \cite{petroni2019language}. By and large, the results reported via such studies have been positive, albeit with significant caveats (see \cite{ettinger2020bert} and \cite{kassner2020negated} for specific criticisms and \cite{pavlick2022semantic} for a general discussion).  

Studies like those above are often not framed overtly in theoretical terms. However, implicitly, they assume a theory of lexical concepts in which meaning is defined via a complex network of associations and inferences \cite{greenberg2005conceptual}. Such theories are a good first step, but contemporary work in psychology has tended to favor a more nuanced picture, in which lexical concepts are embedded in \textit{causal models} (CMs) of the world \cite{keil1989concepts,carey2009origin,sloman2005causal}. These CMs can capture complex inferences about word meaning that have been documented in humans--for example, that a raccoon is still a raccoon even after it has been surgically altered to look and behave like a skunk \cite{keil1989concepts}. Such inferences are not easily explained by theories of concepts that rely on naive association or traditional logical entailment.

In this work, we adopt one such causal model theory of lexical concepts, namely the HIPE theory \cite{Chaigneau:04}, and use it to evaluate whether GPT-3 \cite{brown2020language} and GPT-4 \cite{openai2023gpt4} understand terms referring to basic household objects (specifically, mop, pencil, and whistle). We test GPT-3 and -4 on the stimuli which were used to evaluate humans in the original paper. We find that GPT-3 does not track humans in matching the predictions of HIPE theory about the relative importance of factors determining the concepts tested, even when the experiment is repeated in multiple different variations to guard against a false negative. Contrastingly, we find that GPT-4 tracks humans very well in matching the predictions of HIPE theory on a natural reimplementation of the experiment without introducing experiment variations that would increase the chance of the model’s success. 

Our findings raise important questions about how to evaluate conceptual representations in large language models. In particular, situating our results within a large literature of treating language models as ``psycholinguistic subjects'' \cite{futrell2019neural}, a pertinent question is how to interpret the (increasingly positive) results of LLMs on tests designed to assess humans. If we are hesitant to read success on such tests alone as evidence of ``human-like'' processing (as we the authors are in this case)--what additional testing do we require?

\section{Related Work} 

This work contributes to a large body of work on analyzing LLMs as ``psycholinguistic subjects'' \cite{futrell2019neural} by evaluating their performance on tasks designed to probe human language understanding \cite{marvin2018targeted,warstadt2020blimp,ettinger2020bert}, and more generally to work that uses counterfactual manipulations of model inputs in order to understand model representations \cite{huang-etal-2020-reducing,pmlr-v97-goyal19a}. The phenomena we study relate to past work on ``commonsense'' physical knowledge in LLMs \cite{Bisk:20,Forbes:19,https://doi.org/10.48550/arxiv.2201.07207}, but differs in that we are analyzing an LLM through the lens of a particular, empirically-validated theory about conceptual representations in humans.

\section{The HIPE Theory}
\begin{figure}[ht]
\centering
\begin{tikzpicture}[node distance={30mm}, thick, main/.style = {draw, rectangle, minimum width=2.1cm}] 
\node[main] at (0, 0) (1) {$History$}; 
\node[main] at (0, -0.7) (2) {$Goal$};
\node[main] at (2.75, 0) (3) {$Structure$}; 
\node[main] at (2.75, -0.7) (4) {$Action$};
\node[main] at (5.5, -0.35) (5) {$Outcome$}; 

\draw[->] (1) -- (3);
\draw[->] (2) -- (4);
\draw[->] (3) -- (5);
\draw[->] (4) -- (5);
\end{tikzpicture}
\caption{\label{CM_graph}The CM hypothesized by HIPE theory as underlying human representations of artifacts.}
\end{figure}
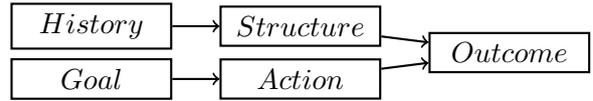

The HIPE theory \citep{Chaigneau:04} aims to explain humans' representations of artifacts (in particular, the work uses the common household objects mops, pencils, and whistles). The HIPE theory posits that humans model an artifact using a causal model (CM) involving the artifact's design history (H), the intentions of relevant agents (I), the object's physical structure (P), and events that occur during its use such as actions taken with it (E). More specifically, the theory posits a particular CM as underlying human reasoning about artifacts (Fig. \ref{CM_graph}). It hypothesizes that the object’s design history and the user's goal are distal causes in the CM, while the object’s physical structure and the user’s actions with respect to it are proximal causes in the CM. Thus, HIPE predicts that, for example, both the physical structure of an object (e.g., having a handle and something absorbent on one end) as well as the reason the object was originally created (e.g., for wiping up water) should affect how appropriate it is to call the object a ``mop'', but that the latter should have a minimal effect when the former is fully specified.

\citet{Chaigneau:04} experimentally confirm that such a CM specifies the structure of human representations of artifacts. They construct scenarios describing the history, structure, goal, and action of the three objects. Each scenario is either a baseline scenario in which all four factors are as one would expect them to be, or a compromised scenario in which exactly one of the factors is altered to a compromised description (Table \ref{prompt_excerpts}). The subjects are then asked to respond to questions about the object's naming (``Is it appropriate to call this object a mop?''), function (``Does this scenario illustrate the function of a mop?''), or causal outcomes (``Is it likely that, as a result of the events described above, John wiped up the water spill?'') using a 1-7 Likert scale. The authors verify that compromising the action has a more pronounced effect than compromising the goal, and likewise that compromising the structure has a more pronounced effect than compromising the design history. This supports the CM's designation of action and structure as proximal causes due to "screening off" \citep{doi:https://doi.org/10.1002/9781118661666.ch31}.

Reimplementing \citet{Chaigneau:04}'s experiment on GPT-3 and -4 is motivated by several factors. First, given that the experiment they use involves text-only stimuli and responses, it can be comparably reimplemented on LLMs with little modification. Second, the CM hypothesized by HIPE theory is intuitive, straightforward, highly general, and relevant for many practical judgements about the physical world. This is unlike, for example, the more subtle theories concerning representation of natural kinds \citep{fosterhanson2021natkinds}. Furthermore, the qualitatively different results we obtain from GPT-3 and -4 are made interesting by the fact that the common household terms studied here (such as ``mop'', ``pencil'', and ``whistle'') seem competently used even by GPT-3 (we document a comprehension test in Section \ref{comp_test} of the appendix to verify this). Thus, our results contribute to teasing apart representational capabilities that are quite similar at face value.  

\begin{figure}
\centering
\begin{tabular}{|p{0.9\columnwidth}|}\hline
\footnotesize
One day Jane wanted to wipe up a water spill on the kitchen floor, but she didn’t have anything to do it with. So she decided to make something. [...] \textit{The object consisted of a bundle of thick cloth attached to a 4-foot long stick}. Later that day, John was looking for something to wipe up a water spill on the kitchen floor. [...] He grabbed the object with the bundle of thick cloth pointing downward and pressed it against the water spill.
\\\hline
\footnotesize
One day Jane wanted to wipe up a water spill on the kitchen floor [...]. \textit{The object consisted of a bundle of plastic bags attached to a 4-foot long stick}. [...] pressed it against the water spill.\\\hline
\end{tabular}
\caption{\label{prompt_excerpts} Examples of scenarios designed to evaluate the HIPE theory. Shown are excerpts from the baseline and compromised structure scenarios with added emphasis. Full prompts are provided in the appendix.}
\end{figure}

\section{Experimental design}

We replicate the crucial first experiment from \citet{Chaigneau:04} on GPT-3 and -4 (model version and hyperparameters are given in Section \ref{hyperparameters_appendix} of the appendix). We investigate the extent to which compromising one of four aspects (goal, action, design history, or physical structure) of a scenario description impacts one of three (causality, function, or naming) outcomes across three artifact types (mop, pencil, or whistle). 

\begin{figure*}[ht!]
\begin{center}
\includegraphics[width=1\linewidth]{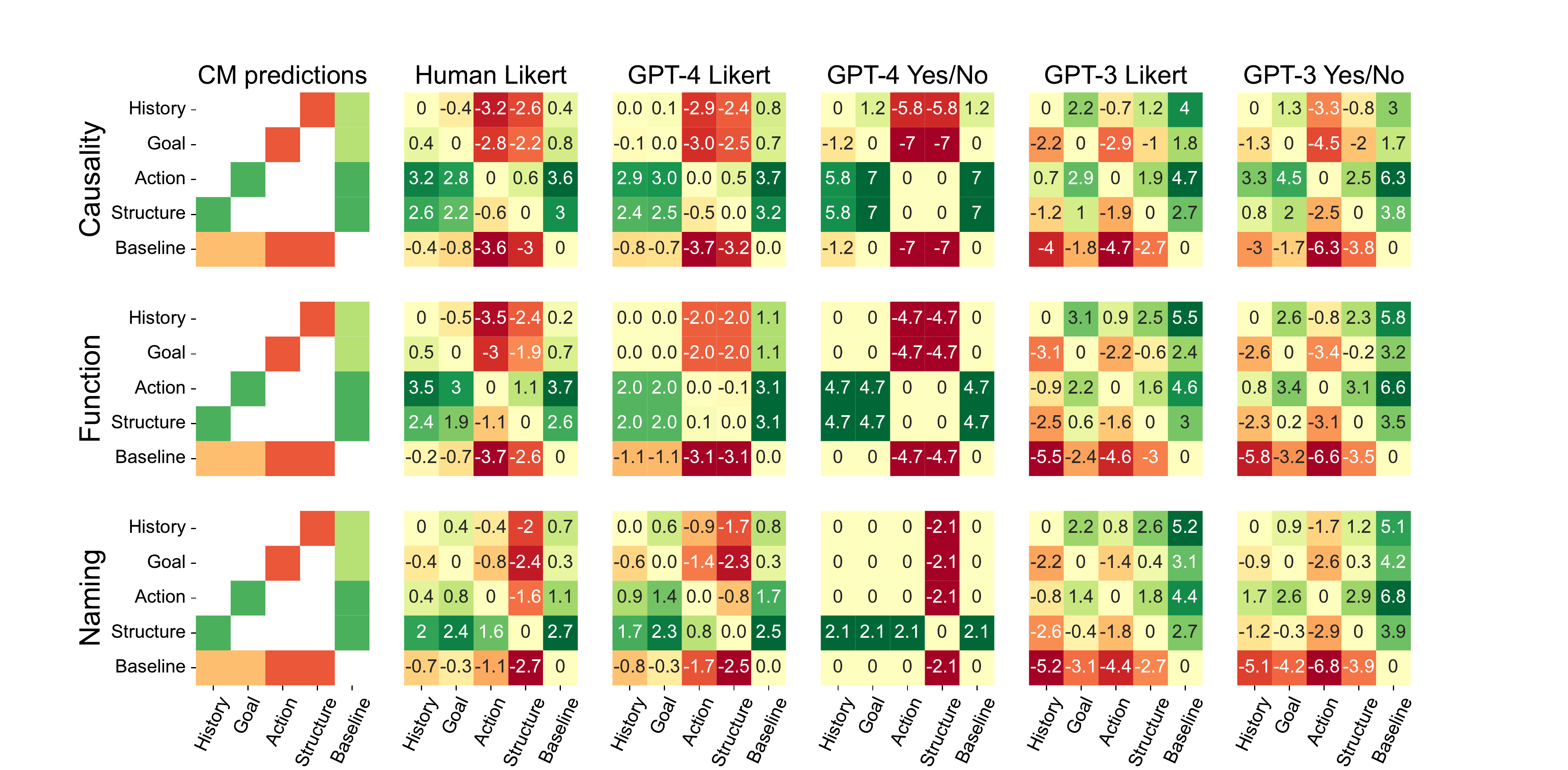}
\end{center}
\caption{\label{yesnoic_heatmaps_paper}Heatmaps showing the pairwise difference between the scores when the factors on the x and y axes are compromised. Column one shows the predictions based on the CM hypothesized by HIPE: compromised scenarios should decrease outcome judgements relative to baseline, compromising action should be more significant than compromising goal, and compromising physical structure should be more significant than compromising design history. Column two shows the human data from \citet{Chaigneau:04}, while columns three to six show the data we obtain from GPT-4 and -3. Columns three and five elicit responses from models using the same Likert scale as in the human experiment, whereas columns four and six use a Yes/No response. Column six directly retrieves generation probabilities from GPT-3, whereas column four averages the frequency of a response across repetitions because access to GPT-4 generation probabilities is not currently available.}
\end{figure*}

First we focus on GPT-3 and consider multiple methods for serving the stimulus to it. Reported in the main body of this paper are the results obtained using a setup that was most faithful to the one humans received, including warm-up trials and the possibility that answers to later questions could be influenced by subjects' exposure to earlier questions. Specifically, the scenarios are served to GPT-3 in a prompt that includes the same guidance that was given to the human participants by \citet{Chaigneau:04}. The first element of the prompt is an introduction consisting of a description of the experiment and several demonstration responses. After the introduction the uncompromised scenario is presented and the Yes/No response is recorded.  The compromised scenarios are then presented in random order. For each new scenario, we allow GPT-3 access to its entire response history as part of the prompt (since humans would have memory of their own past responses). The prompts we construct can thus be seen as a type of ``in context learning''. That is, when GPT-3 is generating a Yes/No reply for the $k+1$ scenario, it receives the introductory instructions and examples as well as the sequence of the first $k$ scenarios and its own generated replies to those scenarios as part of the prompt. We record the probability that the model assigns to ``yes'' or versions thereof in response to each scenario. To control for possible effects from the order in which the compromised prompts are presented, three sets of results are averaged. Figure \ref{prompt_excerpts} shows excerpts from the prompts provided to GPT-3. 

We also consider a further five variations on the above prompting design, which differ in how the scenarios are presented to GPT-3 and the manner in which a response is recorded. However, these alternative variations yield a lower Spearman correlation between the human and GPT-3 results (ranging between 0.28 and 0.5, versus 0.81 for the version presented here - see Figure \ref{experiment_corrs} of the appendix) and are less true to the original experiment. The negative result we will report for GPT-3 is strengthened by these numerous attempts to adapt the stimulus to it. By contrast, we will report positive results for GPT-4 despite only implementing the most faithful re-construction of the original experiment on it.

We reimplement the final version of the above experiment on GPT-4 following its release. However, since OpenAI does not support the retrieval of probabilities associated with tokens generated by GPT-4, we instead repeat each question ten times and calculate the probability that the model generates a response including ``yes'' or versions thereof. The ten responses are split across two runs of five to control for the particular random order in which questions are presented. We report results in Figure \ref{yesnoic_heatmaps_paper} using the system message ``You are a helpful assistant'' and include further results in Figure \ref{gpt4_pw_heatmap} of the appendix with the system message ``You are a helpful assistant with an excellent understanding of the physical world'' (this slightly increases the correlation between the results from GPT-4 and human subjects, but is not necessary for observing a positive result). For parity we redo this experiment on GPT-3 with repeated output generation instead of the direct retrieval of output probabilities. We find a Spearman correlation of 0.96 between the experiments run on GPT-3 with and without direct probability retrieval, giving us confidence that the results obtained from GPT-4 without direct access to generation probabilities are comparable to what would be obtained with direct access. Because we find the results from GPT-4 to be more binary when using the Yes/No response as compared to the human data which was collected using a Likert scale, we re-implement the experiment with GPT-4 using the ``Helpful assistant'' system prompt and a Likert response. For parity we do the same with GPT-3.

We further verify that GPT-4 responds in the expected way to compromising distal factors by using a Likert scale re-implementation of Experiment 2 from \citet{Chaigneau:04}, which tests a cumulative effect of compromising both distal factors while leaving the proximal factors unchanged. After providing the model with the same introductory prompt as above, we provide it with the baseline scenario, elicit its response on a Likert scale, accumulate this response, and then provide it successively with the next three scenarios in which one or both of function and history are compromised. We repeat this experiment for all three objects, running a given question / object combination twice with five responses collected from the model each time. We omit the naming question from this experiment for comparability with \citet{Chaigneau:04}, who do the same.

\section{Results}

\subsection{Experiment 1}

Figure \ref{yesnoic_heatmaps_paper} shows the predictions made according to the CM hypothesized by HIPE, the results obtained by \citet{Chaigneau:04} on human subjects, and the results obtained by us on GPT-4 and -3. The CM predicts that history should have a less significant effect on outcome judgements than structure, and that goal should have a less significant effect than action. This corresponds to the diagonal of two green and two orange boxes in the CM predictions column of the figure. Furthermore, the CM predicts that compromising any factor should have a negative effect on outcome judgements relative to baseline, but with distal factors yielding a smaller negative effect than proximal factors. This corresponds to the orange horizontal and green vertical bars in the CM predictions column, lightening towards the left and top respectively due to the weaker effect of the distal factors.

The results from human subjects abide neatly by these predictions in the case of causality and function judgements. We observe a clear red box of four cells towards the top right, mirrored by a green box towards the bottom left. This subsumes the green/orange diagonal of the CM prediction, and corresponds to the stronger result of a larger effect of each proximal factor than both distal factors, rather than only a weaker result of structure being more significant than history and action being more significant than goal. We also see a green vertical bar on the right and a corresponding red bar on the bottom, lightening at the top and left respectively. This corresponds to every factor making a negative difference relative to baseline, with distal factors mattering less than proximal factors. 

In the naming case, we see a somewhat different pattern in the human data than predicted by the CM. As the CM predicts, we observe a green column to the right that lightens in the top half, indicating that compromising any factor compromises the outcome judgement, but that compromising proximal factors compromises the outcome judgement to a greater extent. However, a prominent red column in the fourth position mirrored by a green horizontal fourth row corresponds to a larger negative effect on the outcome when compromising structure than when compromising other features. This is intuitively reasonable: for example, using a bowl as a spoon does less to make it no longer be a bowl than flattening it does. 

The results from GPT-4 bear a striking resemblance to the human data. In the causality and function heatmaps, we see a green vertical on the right and a red horizontal on the bottom, lightening towards the top right and bottom left respectively. We also see strong red boxes in the top right mirrored by green boxes in the bottom left. In the naming case, we see the same strong red column and green row appear in the fourth positions, corresponding to a dominating effect of compromising structure on the outcome compared to the effect of compromising other factors.

Overall there is a 0.88 Spearman correlation between the GPT-4 and human data with a Yes/No response (column four of Figure \ref{yesnoic_heatmaps_paper}). The results from GPT-4 collected with a Yes/No response are more binary than the results from human subjects that were collected using a Likert scale response. In particular, the human data shows some effect of compromising distal factors thus not demonstrating full screening off of the distal factors by the proximal ones, while the Yes/No-response data from GPT-4 does not show this property. This is due to the difference in response modality, and we verify that re-implementing the experiment on GPT-4 with a Likert-scale response (column three of Figure \ref{yesnoic_heatmaps_paper}) eliminates this effect and increases the Spearman correlation with the human data from 0.88 to 0.92. 

The results from GPT-3 are much less consistent with the predictions based on the CM and with the human data. First we focus on the results elicited using the same Likert scale from the human experiment (column five of Figure \ref{yesnoic_heatmaps_paper}). Across all three questions, we see that GPT-3 (like humans) consistently considers the compromised scenarios as less consistent with the concept than the baseline scenario. However, when comparing the effect of history to structure and the effect of goal to action across the three questions, GPT-3’s responses only agree with the CM predictions in 3 out of 6 cases. Moreover, the Spearman correlation with the human data is only 0.67. However, these results may be due simply to the failure of GPT-3 to competently use a Likert scale. Indeed, after initially experimenting with a Likert scale on GPT-3, this approach was abandoned due to evidence that the model was not competently using such a response format. The results from a Likert scale experiment on GPT-3 are included here primarily for comparability with GPT-4, as the latter model appears to competently use the scale and exhibits the closest similarity to the human data when using this setup from the original human experiment. 

When replacing the Likert scale with a simpler to use Yes/No response (column six of Figure \ref{yesnoic_heatmaps_paper}), the results from GPT-3 correlate more closely with the human data (Spearman correlation = 0.81) but differ in key respects. When comparing the effect of history to structure and the effect of goal to action across the three questions, GPT-3’s responses only agree with the CM predictions in 4 out of 6 cases. While both the human and GPT-4 data show clear red boxes in the upper right corresponding to a much more significant effect of proximal than distal causes (with this effect becoming more pronounced when binary data are collected from GPT-4), the data from GPT-3 show a more mixed spread without clear patterns emerging when switching to binary Yes/No responses. Furthermore, the human data show a pattern of high Spearman correlation between causality and function questions (0.99) with a much lower correlation between those questions and the naming one (0.64 causal / naming, 0.60 function / naming). Similarly, the GPT-4 data show correlations of 0.95 causal / function, 0.58 causal / naming, and 0.54 function / naming in the Yes/No response version. By contrast, in the GPT-3 Yes/No response data the correlation between questions is high in all comparisons (all pairwise correlations $\geq$ 0.93). Additionally, while in the naming question we see a very strong effect of structure compared to all other factors in the data from humans and GPT-4, we see a stronger effect from action in the GPT-3 data (this can be seen in the redder fourth columns of the naming plots from the human and GPT-4 data, compared to the redder third column in the equivalent plot from the GPT-3 data). These discrepancies suggest non-trivial differences between how human subjects and GPT-4 on the one hand and GPT-3 on the other process these questions.

\subsection{Experiment 2}

Experiment 1 primarily tests that the subject exhibits the screening off of distal factors: i.e., when distal factors are compromised but the proximal factors that are hypothesized to mediate their effect are left unchanged, the compromising effect of the distal factors should be largely masked. 

Nevertheless, one should expect compromising the distal factors to have some effect on outcome judgements and indeed this is observed in Experiment 1. Experiment 2 further investigates the effect of compromising distal factors by verifying that compromising each distal factor independently results in a lowered outcome judgement relative to baseline and that compromising both of these factors together results in an even lower outcome judgement. Following \citet{Chaigneau:04} for comparability, we present results from GPT-4 that average function and causal outcome judgements across the three object types. 

\begin{figure}[ht!]
\begin{center}
\includegraphics[width=1\linewidth]{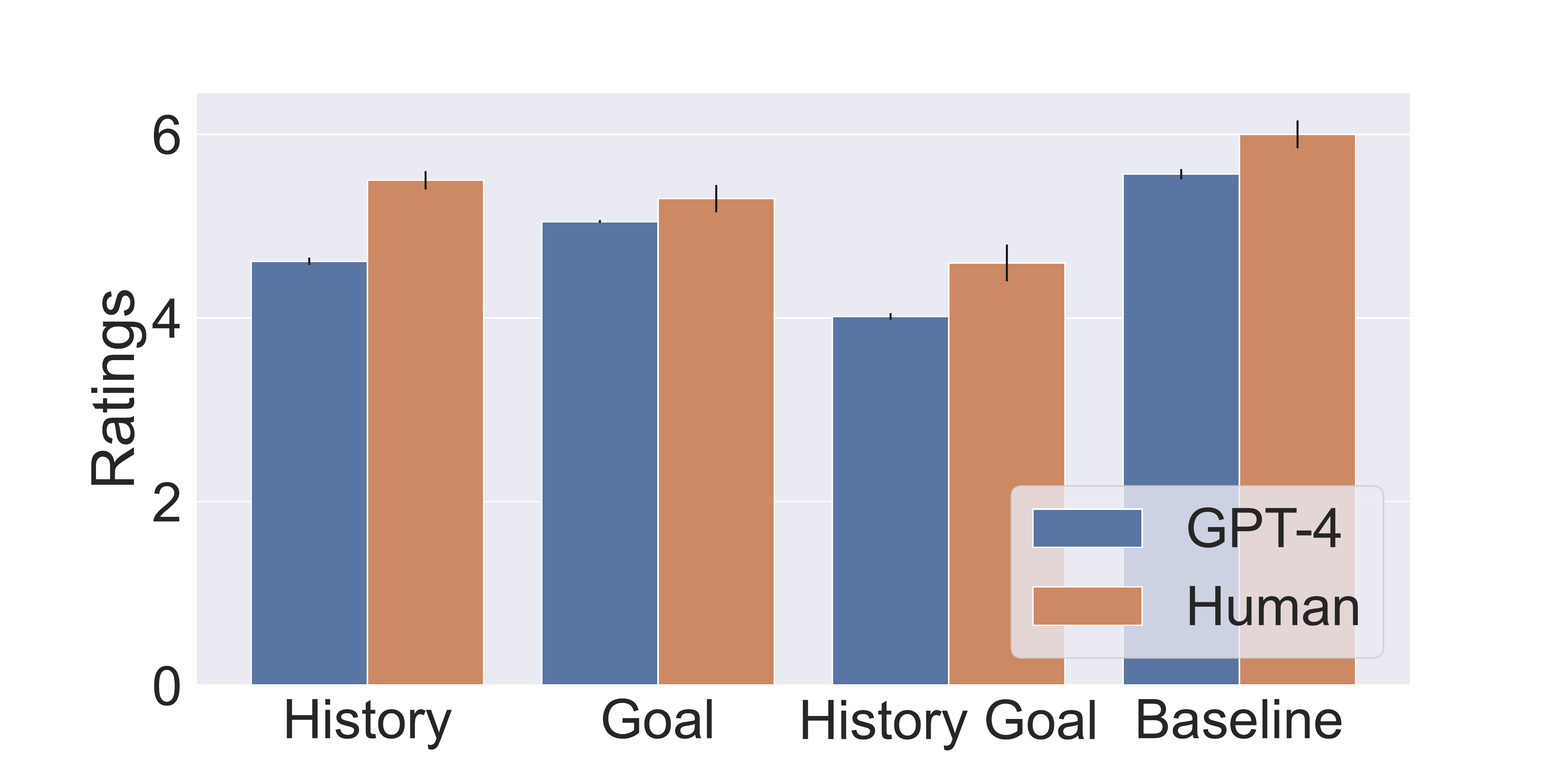}
\end{center}
\caption{\label{exp2_barplot}Human and GPT-4 data collected in Experiment 2. Human data are from \citet{Chaigneau:04}. The chart shows the subject's mean rating, averaged across function and causality for all three objects, when the factor or factors shown on the x-axis are compromised. Error bars are the Standard Error of the Mean.}
\end{figure}

As can be observed in Figure \ref{exp2_barplot}, the data collected from GPT-4 and human subjects follow the same trend. In particular, compromising history, goal, or both together each yield a lower outcome rating than baseline. Moreover, compromising both of these distal factors together yields a lower outcome rating than compromising either of them independently. \citet{Chaigneau:04} find that these five pairwise comparisons are statistically significant in the human data, and we find that the same holds in the data collected from GPT-4. Further information is shown in Figure \ref{exp2_table}.

\begin{figure}
\begin{center}
\begin{tabular}{ c | c c }
 Comparison & t value & p value \\ 
 \hline
 History $<$ Baseline & 13.7 & $\le$ 0.0005 \\  
 Goal $<$ Baseline & 8.9 & $\le$ 0.0005 \\ 
 History Goal $<$ Baseline & 23.5 & $\le$ 0.0005 \\
 History $>$ History Goal & 11.2 & $\le$ 0.0005 \\
 Goal $>$ History Goal & 27.2 & $\le$ 0.0005 \\
\end{tabular}
\end{center}
\caption{\label{exp2_table}Table showing details from statistical testing of the difference between mean response values across the questions of Experiment 2. \citet{Chaigneau:04} conduct the same comparisons using the human data, finding that the expected differences are significant to at least the p $\le$ 0.01 level. We find that the same differences are significant to at the least the p $\le$ 0.0005 level. Our data for the statistical testing are the responses from GPT-4 shown in Figure \ref{exp2_barplot}.}
\end{figure}

While the absolute magnitude of the responses differ between humans and GPT-4, this may be due to the different calibration of human subjects and GPT-4 when using the Likert scale. Human subjects are prompted with ``warm-up'' questions using the Likert scale, which they are encouraged to discuss with each other. We have access to these questions but not records of human responses, and so we provide GPT-4 with these questions accompanied with our own responses to them as part of its prompt. This successfully primes GPT-4 to use the scale meaningfully, but it should not be expected to be calibrated to the original human subjects. 

\section{Discussion}

Our results show a similarity between the responses of GPT-4 and human subjects in this experiment. By contrast, our results show a marked difference between the responses from human subjects and GPT-4 on the one hand and GPT-3 on the other, suggesting a qualitative difference between GPT-3 and -4 in how they represent such artifacts. 

At the highest level, we interpret these results as speaking to the need for a broad and rigorous discussion about evaluation in the modern age of LLMs. Recent years have relied increasingly on tests from cognitive science and psycholinguistics as a source of more rigorous, more controlled, and more hypothesis-driven evaluations of language models \cite{blackboxnlp-2022-blackboxnlp}. Such experiments have been primarily fruitful in the context of two types of arguments. First, they have produced insightful negative results (e.g., \citet{ettinger2020bert}). In such cases, models' failure on psycholinguistic tests can be taken as evidence that the models probably lack at least \textit{some} aspect of whatever mechanism humans use to perform the same tasks. Second, such tests have produced insightful positive results (e.g., \citet{linzen-etal-2016-assessing}). For example, models' success has been used specifically to counter learnability or ``poverty of the stimulus'' arguments, and thus to question the usefulness of specific diagnostic tests. That is, if some behavior is assumed to require a given capacity, and a model that is known to lack that capacity nonetheless produces that behavior, then a different test is needed to diagnose the capacity of interest. 

The present study may best be viewed as an instance of the latter. Theories like HIPE are generally assumed to be tests of causal models which presuppose that agents' representations are grounded in the physical and goal-oriented world. If models with access only to text (or at most text and images) are presumed to lack this grounding, then the success of models on this task may suggest that the human results on HIPE tests are not \textit{necessarily} diagnostic of such grounding. Thus, further tests must be developed to determine what representations underlie models' (and humans') behavior in these contexts.

Importantly, caution should be exercised in interpreting positive results on psychological tests as diagnostic of ``human-like'' or even ``human-level'' processing. Only in some cases are positive results in such tests clearly interpretable, such as against a backdrop of a clear learnability argument (i.e., a claim about some capacity that the model being studied is known \textit{a priori} not to possess). However, the likelihood of increased positive results in the age of LLMs could lead to psycholinguistic tests being hastily viewed as diagnostic of human-like processing. Thus, we raise questions about what role such tests should play in future evaluations of models' representations. In particular, if success on behavioral tests alone is not a sufficient test of competence, what is? 



\section*{Limitations}

We use the same materials as \citet{Chaigneau:04} and aim to preserve comparability with the data they collect from human subjects. Two limitations arise from this. First, we consider only the three artifacts from the original study (mops, pencils, and whistles) and average results across them. Results could differ if more objects were included. However, comparable data from human subjects do not appear to exist for a broader class of objects.

Second, we use the scenario descriptions from the original experiment and therefore are vulnerable to peculiarities in the way these scenarios were written. On the whole the materials from the original study are of a high quality. Nevertheless, some issues are present. For example, consider the causal question for the whistle object in the compromised action scenario. The whistle object consists of a shell with its tip broken off. In the compromised action scenario, the character attempts to call his dog (which is trained to respond to a high-pitched sound) by whispering the dog’s name into the shell instead of blowing into the opening at its tip. The compromised action is meant to not result in the expected causal outcome, in this case the dog being summoned. However, whispering a dog’s name into a shell is quite likely to result in the dog being called. Indeed, GPT-3 responds that the dog is likely to be summoned, and this data is included in what we report. We investigate this further by altering the compromised action to whispering a random word into the shell, which unambiguously would not result in the dog being summoned. As expected, GPT-3 changes its response to indicate that the dog would not be called. While the original human subjects are exposed to the same deficiencies in the scenario descriptions that the LLMs are, it is possible that humans and the LLMs are differentially sensitive to such shortcomings.  

Last, our research is limited in that we only investigate terms referring to objects, and we only investigate English terms.

\section*{Ethics Statement}
Data were obtained from API access to Open AI’s GPT-3 and -4 models, as allowed under their Terms of Use. The Terms of Use do not specify that research purposes are not consistent with the intended use of the model. No experiments involving human subjects were performed, as all human data were reported from \citet{Chaigneau:04}. Because the paper is a study of an existing model, no concerns regarding potential harmful misuse arise. The experiment involved a modest amount of computation, as reflected in moderate billing charges from use of Open AI’s API. Therefore, the environmental impact of the work is low. 

\bibliography{anthology,custom}
\bibliographystyle{acl_natbib}

\appendix

\section{Further experiment details}
\label{sec:further_exp_details}

\subsection{Model and hyperparameter settings}
\label{hyperparameters_appendix}

We run our experiments on GPT-3 with the following parameters: 

Engine: text-davinci-002

Temperature:0.7

Max tokens:5

Top p:1

Frequency penalty:0

Presence penalty:0
\newline
\newline
We run our experiments on GPT-4 with the following parameters: 

Engine: gpt-4-0314 (i.e., the March 14 2023 snapshot of GPT-4)

Temperature:0.7

Max tokens:5

Top p:1

Frequency penalty:0

Presence penalty:0

\subsection{Testing GPT-3's comprehension of the domain}
\label{comp_test}

We run a simple experiment in which we present GPT-3 with 5 questions about each object (mop, pencil, and whistle). We verify that GPT-3 seems able to easily generate coherent text about these objects in ways that are relevant for the main experiment we conduct. This makes the negative results for that model in our main experiment more surprising. We use the general model and hyperparameter settings from above, but we increase max tokens to 256. 

The model correctly responds to the question in 13 out of 15 cases. In 1 case it is correct but does not address the question, and in 1 case its response is incorrect assuming the question is construed literally. Overall, these are positive results.

Mop

1.	What is the function of a mop? 

A mop is a household cleaning tool that is used to clean floors.

Evaluation: Correct

2.	How do you use a mop? 

To use a mop, first dip the mop head in a bucket of soapy water. Next, wring out the mop head so that it is only slightly damp. Finally, mop the floor in a back and forth motion.

Evaluation: Correct

3.	Would a mop still work if it was made with unabsorbent materials?

A mop would not work if it were made with unabsorbent materials.

Evaluation: Correct

4.	If you press the end of a mop against water, will it soak it up? 

The mop will soak up water if it is pressed against it.

Evaluation: Correct

5.	Can you use a mop to paint a pretty picture?

No, you cannot use a mop to paint a pretty picture.

Evaluation: Correct

Pencil

1.	What is the function of a pencil? 

A pencil is a tool used for writing and drawing.

Evaluation: Correct

2.	How do you use a pencil?

A pencil is used by holding it in your hand and using your fingers to move it across a piece of paper.

Evaluation: Correct

3.	Would a pencil still work if it was a polished stick without graphite in the middle?

No, a pencil needs graphite in the middle in order to work.

Evaluation: Correct

4.	If you drag the tip of a pencil across a piece of paper will it leave a mark? 

Yes, it will leave a mark.

Evaluation: Correct

5.	Can you use a pencil to nail two pieces of wood together?

No

Evaluation: Correct

Whistle

1.	What is the function of a whistle? 

The function of a whistle is to produce a loud, shrill sound that can be heard from a long distance.

Evaluation: Correct

2.	How do you use a whistle? 

A whistle can be used for many things, such as to signal for help, to get someone's attention, or to warn of danger.

Evaluation: Correct but does not address the question

3.	Would a whistle still work if it was blocked?

A whistle would not work if it was blocked.

Evaluation: Correct

4.	If you blow into a whistle and there’s a dog nearby will the dog respond?

There's no certain answer to this question since it depends on the dog's training, personality, and level of responsiveness. However, it's possible that the dog may come over to investigate the source of the noise if it is curious or believes that the person blowing the whistle is calling for it.

Evaluation: Correct

5.	Can you use a whistle to speak loudly?

Yes

Evaluation: Incorrect if the question is interpreted literally

\subsection{Prompt materials reused across experiments}

Here we include the prompt materials that are reused across experiments. These materials are included in prompts served to GPT-3 in different ways the various versions of the experiment documented below. 

Mop object, uncompromised scenario: 

One day Jane wanted to wipe up a water spill on the kitchen floor, but she didn’t have anything to do it with. So she decided to make something. She looked around the house for things that would allow her to make an object for wiping up a water spill on the kitchen floor. She gathered all the materials and made it. When she finished, she left it in the kitchen so she could use it later. The object consisted of a bundle of thick cloth attached to a 4-foot long stick. Later that day, John was looking for something to wipe up a water spill on the kitchen floor. He saw the object that Jane had made and thought that it would be good for wiping up a water spill on the kitchen floor. He grabbed the object with the bundle of thick cloth pointing downward and pressed it against the water spill.

Mop object, compromised history scenario: 

One day Jane was cleaning the attic. She picked up a bunch of useless things and put them all inside a big cardboard box. Because the box was overflowing, she used a long stick to shove things down. As she did this, something became attached to the stick. Then, Jane carried the box downstairs. She didn’t notice that as she did this, the stick and the thing that was attached to it fell together, as a single object, to the floor. The object consisted of a bundle of thick cloth attached to a 4-foot long stick. Later that day, John was looking for something to wipe up a water spill on the kitchen floor. He saw the object that Jane had made and thought that it would be good for wiping up a water spill on the kitchen floor. He grabbed the object with the bundle of thick cloth pointing downward and pressed it against the water spill.

Mop object, compromised structure scenario: 

One day Jane wanted to wipe up a water spill on the kitchen floor, but she didn’t have anything to do it with. So she decided to make something. She looked around the house for things that would allow her to make an object for wiping up a water spill on the kitchen floor. She gathered all the materials and made it. When she finished, she left it in the kitchen so she could use it later. The object consisted of a bundle of plastic bags attached to a 4-foot long stick. Later that day, John was looking for something to wipe up a water spill on the kitchen floor. He saw the object that Jane had made and thought that it would be good for wiping up a water spill on the kitchen floor. He grabbed the object with the bundle of plastic bags pointing downward and pressed it against the water spill.

Mop object, compromised goal scenario: 

One day Jane wanted to wipe up a water spill on the kitchen floor, but she didn’t have anything to do it with. So she decided to make something. She looked around the house for things that would allow her to make an object for wiping up a water spill on the kitchen floor. She gathered all the materials and made it. When she finished, she left it in the kitchen so she could use it later. The object consisted of a bundle of thick cloth attached to a 4-foot long stick. Later that day, John was in the kitchen looking for something to eat. He was distracted as he looked for something and inadvertently grabbed the object that Jane had left in the kitchen. He grabbed the object with the bundle of thick cloth pointing downward and pressed it against the water spill.

Mop object, compromised action scenario: 

One day Jane wanted to wipe up a water spill on the kitchen floor, but she didn’t have anything to do it with. So she decided to make something. She looked around the house for things that would allow her to make an object for wiping up a water spill on the kitchen floor. She gathered all the materials and made it. When she finished, she left it in the kitchen so she could use it later. The object consisted of a bundle of thick cloth attached to a 4-foot long stick. Later that day, John was looking for something to wipe up a water spill on the kitchen floor. He saw the object that Jane had made and thought that it would be good for wiping up a water spill on the kitchen floor. He grabbed the object with the bundle of thick cloth pointing upward instead of downward, and pressed the bare wood end against the water spill. 

Pencil object, uncompromised scenario: 

One day Jane wanted to draw lines on a white sheet of paper, but she didn’t have anything to do it with. So she decided to make something. She looked around the house for things that would allow her to make an object for drawing lines on a white sheet of paper. She gathered all the materials and made it. When she finished, she left it on a table so she could use it later. The object consisted of a slender wooden stick, approximately 3 inches in length, which had been lightly burned. Later that day, John was looking for something to draw lines on a white sheet of paper. He saw the object that Jane had made and thought that it would be good for drawing lines on a white sheet of paper. He grabbed the object and pressed its tip against the white sheet of paper while moving his hand in different directions.

Pencil object, compromised history scenario: 

One day Jane noticed that the fireplace needed to be cleaned. She piled up the ashes, half-burned logs, and sticks and carefully transferred everything into an ash bucket. She didn’t notice that as she did this, one object fell on the floor. The object consisted of a slender wooden stick, approximately 3 inches in length, which had been lightly burned. Later that day, John was looking for something to draw lines on a white sheet of paper. He saw the object that Jane had made and thought that it would be good for drawing lines on a white sheet of paper. He grabbed the object and pressed its tip against the white sheet of paper while moving his hand in different directions.

Pencil object, compromised structure scenario: 

One day Jane wanted to draw lines on a white sheet of paper, but she didn’t have anything to do it with. So she decided to make something. She looked around the house for things that would allow her to make an object for drawing lines on a white sheet of paper. She gathered all the materials and made it. When she finished, she left it on a table so she could use it later. The object consisted of a slender wooden stick, approximately 3 inches in length, that had been polished with sandpaper. Later that day, John was looking for something to draw lines on a white sheet of paper. He saw the object that Jane had made and thought that it would be good for drawing lines on a white sheet of paper. He grabbed the object and pressed its tip against the white sheet of paper while moving his hand in different directions.

Pencil object, compromised goal scenario: 

One day Jane wanted to draw lines on a white sheet of paper, but she didn’t have anything to do it with. So she decided to make something. She looked around the house for things that would allow her to make an object for drawing lines on a white sheet of paper. She gathered all the materials and made it. When she finished, she left it on a table so she could use it later. The object consisted of a slender wooden stick, approximately 3 inches in length, which had been lightly burned. Later that day, John was sitting at the table while eating his lunch. He was distracted as he munched and inadvertently grabbed the object that Jane had left on the table. He grabbed the object and pressed its tip against the white sheet of paper while moving his hand in different directions.

Pencil object, compromised action scenario: 

One day Jane wanted to draw lines on a white sheet of paper, but she didn’t have anything to do it with. So she decided to make something. She looked around the house for things that would allow her to make an object for drawing lines on a white sheet of paper. She gathered all the materials and made it. When she finished, she left it on a table so she could use it later. The object consisted of a slender wooden stick, approximately 3 inches in length, which had been lightly burned. Later that day, John was looking for something to draw lines on a white sheet of paper. He saw the object that Jane had made and thought that it would be good for drawing lines on a white sheet of paper. He grabbed the object and waved it in front of the white piece of paper without ever touching it.

Whistle object, uncompromised scenario: 

One day Jane wanted to call her dog (who was out in the garden and was trained to answer to a high-pitch sound), but she didn’t have anything to do it with. So she decided to make something. She looked around the house for things that would allow her to make an object for calling her dog. She gathered all the materials and made it. When she finished, she left it on a table so she could use it later. The object was a conical sea shell that now had its tip broken. Later that day, John was looking for something to call his dog with. He saw the object that Jane had made and thought that it would be good for calling his dog. He grabbed the object, put its tip in his mouth, and blew.

Whistle object, compromised history scenario: 

One day Jane wanted to clean up her desk. She reviewed different documents and objects that were on her desk and began to put all unwanted items in a cardboard box. Because she wasn’t careful when throwing objects into the box, the tip of one of the objects she discarded broke. The object was a conical sea shell that now had its tip broken. Later that day, John was looking for something to call his dog with. He saw the object that Jane had made and thought that it would be good for calling his dog. He grabbed the object, put its tip in his mouth, and blew.

Whistle object, compromised structure scenario: 

One day Jane wanted to call her dog (who was out in the garden and was trained to answer to a high-pitch sound), but she didn’t have anything to do it with. So she decided to make something. She looked around the house for things that would allow her to make an object for calling her dog. She gathered all the materials and made it. When she finished, she left it on a table so she could use it later. The object was a conical sea shell that now had its tip broken and replaced with a solid piece of plastic resin that completely blocked the opening. Later that day, John was looking for something to call his dog with. He saw the object that Jane had made and thought that it would be good for calling his dog. He grabbed the object, put its tip in his mouth, and blew.

Whistle object, compromised goal scenario: 

One day Jane wanted to call her dog (who was out in the garden and was trained to answer to a high-pitch sound), but she didn’t have anything to do it with. So she decided to make something. She looked around the house for things that would allow her to make an object for calling her dog. She gathered all the materials and made it. When she finished, she left it on a table so she could use it later. The object was a conical sea shell that now had its tip broken. Later that day, John was searching on the table for something to play with. He was distracted as he looked for something and inadvertently grabbed the sea shell. He grabbed the object, put its tip in his mouth, and blew.

Whistle object, compromised action scenario: 

One day Jane wanted to call her dog (who was out in the garden and was trained to answer to a high-pitch sound), but she didn’t have anything to do it with. So she decided to make something. She looked around the house for things that would allow her to make an object for calling her dog. She gathered all the materials and made it. When she finished, she left it on a table so she could use it later. The object was a conical sea shell that now had its tip broken. Later that day, John was looking for something to call his dog with. He saw the object that Jane had made and thought that it would be good for calling his dog. He grabbed the object, put his mouth near the wider opening, and whispered his dog’s name.

\subsection{Experiment A: Likert scale}

In this experiment, the prompt presented to the LLM consists of a single scenario followed by a sentence requesting a Likert scale response. An example of the sentence following a scenario is as follows: “On an inclusive rating scale from 1 to 7, with 1 representing very unlikely and 7 representing very likely, how likely would it be that, as a result of the events described above, John wiped up the water spill? The rating is:”. The numerical response is recorded and averaged across five repetitions. The question is replaced as appropriate to match the object type and the causal, function, or naming condition. The example shown here is the causal question for the mop object. 

Experimenting with this setup showed a propensity of the model to provide somewhat undifferentiated responses of 3, 4, and 5. Testing also showed deficiencies with the model's ability to use the scale correctly, as responses were significantly different when asked to use a 1-20 or a 1-100 scale. As a result, we only implemented this experiment for the mop object before moving onto alternative experiment designs. 

Prompts were presented to GPT-3 in the following format: 

``[Scenario]. On an inclusive rating scale from 1 to 7, with 1 representing very unlikely and 7 representing very likely, [Question]? The rating is:''

The scenario is one of the five mop scenarios documented above in the materials used across experiments. 

The question is one of the following: 

Casual: how likely would it be that, as a result of the events described above, John wiped up the water spill? 

Function: how well does this scenario illustrate the function of a mop?

Naming: how appropriate is it to call this object a mop?

\subsection{Experiment B: Percent yes, Yes/No}

In this experiment, the prompt presented to the LLM consists of an introduction, followed by a single scenario and then a request for a Yes/No output. The version of the introduction for the naming question is as follows: 

``Instruction: respond to the scenario with either yes or no.

Here are two example responses.

It is appropriate to call this object a wallet: yes

It is appropriate to call this object a hammer: no

Now provide a similar response based on the scenario below.''

This is followed by a single scenario description as documented in the materials reused across experiments, and then a request for output such as: 

Causal: ``It is likely that, as a result of the events described above, John wiped up the water spill:''

Function: ``This scenario illustrates the function of a mop:''

Naming: ``It is appropriate to call this object a mop:''

The recorded output is the probability that the model assigns to ``yes'' or versions thereof (such as ``YES'').

Each prompt served to GPT-3 includes only the introductory segment, a single scenario, and a single question. 

\subsection{Experiment C: Percent yes, Yes/No/Maybe}

This experiment is much like experiment B, except that now we prompt the model to respond with yes, no, or maybe. 

The introductory text is therefore changed to the following: 

``Instruction: respond to the scenario with yes, no, or maybe.

Here are three example responses.

It is appropriate to call this object a wallet: yes

It is appropriate to call this object a hammer: no

It is appropriate to call this object a mug: maybe

Now provide a similar response based on the scenario below.''

Again we record the probability that the model assigns to ``yes'' or versions thereof. The hypothesis was that a less binary prompting might coax the model to more subtly distinguish between prompt versions.

\subsection{Experiment D: Percent yes, Yes/No, shortened prompts}

This experiment is much like experiment B, except that now we prompt the model with maximally shortened versions of the scenarios (approximately 50 words each, as opposed to the original 150). Again we record the probability that the model assigns to “yes” or versions thereof as the output. The hypothesis was that shorter prompts may aid the model’s comprehension of the scenarios. 

The shortened scenario descriptions are listed below. 

Mop object, shortened uncompromised scenario: 

Jane wanted to wipe up water, so she made something. The object was a bundle of cloth attached to a long stick. John wanted to wipe up water, so he picked up the object. He pressed the end with the bundle against the water spill. 

Mop object, shortened compromised history scenario: 

Jane dropped two things which became attached to each other as one object. The object was a bundle of cloth attached to a long stick. John wanted to wipe up water, so he picked up the object. He pressed the end with the bundle against the water spill. 

Mop object, shortened compromised structure scenario: 

Jane wanted to wipe up water, so she made something. The object was some plastic bags attached to a long stick. John wanted to wipe up water, so he picked up the object. He pressed the end with the plastic bags against the water spill. 

Mop object, shortened compromised goal scenario: 

Jane wanted to wipe up water, so she made something. The object was a bundle of cloth attached to a long stick. John was looking for something to eat and inadvertently picked up the object. He pressed the end with the bundle against the water spill. 

Mop object, shortened compromised action scenario: 

Jane wanted to wipe up water, so she made something. The object was a bundle of cloth attached to a long stick. John wanted to wipe up water, so he picked up the object. He pressed the bare end of the stick, without the bundle, against the water spill. 

Pencil object, shortened uncompromised scenario: 

Jane wanted to draw lines on paper, so she made something. The object was a slender burnt wooden stick. John wanted to draw lines on paper, so he picked up the object. He pressed its tip against the paper while moving his hand.

Pencil object, shortened compromised history scenario: 

Jane dropped an object from the fireplace. The object was a slender burnt wooden stick. John wanted to draw lines on paper, so he picked up the object. He pressed its tip against the paper while moving his hand. 

Pencil object, shortened compromised structure scenario: 

Jane wanted to draw lines on paper, so she made something. The object was a slender wooden stick that had been polished with sandpaper. John wanted to draw lines on paper, so he picked up the object. He pressed its tip against the paper while moving his hand. 

Pencil object, shortened compromised goal scenario: 

Jane wanted to draw lines on paper, so she made something. The object was a slender burnt wooden stick. John was eating his lunch and inadvertently picked up the object. He pressed its tip against the paper while moving his hand. 

Pencil object, shortened compromised action scenario: 

Jane wanted to draw lines on paper, so she made something. The object was a slender burnt wooden stick. John wanted to draw lines on paper, so he picked up the object. He moved it around in front of the paper without ever touching it.

Whistle object, shortened uncompromised scenario: 

Jane wanted to call the dog, so she made something. The object was a shell that now had its tip broken. John wanted to call the dog, so he picked up the object. He put its tip in his mouth and blew. 

Whistle object, shortened compromised history scenario: 

Jane dropped an object, which broke. The object was a shell that now had its tip broken. John wanted to call the dog, so he picked up the object. He put its tip in his mouth and blew. 

Whistle object, shortened compromised structure scenario: 

Jane wanted to call the dog, so she made something. The object was a shell that now had its tip broken, with the opening completely blocked by plastic. John wanted to call the dog, so he picked up the object. He put its tip in his mouth and blew. 

Whistle object, shortened compromised goal scenario: 

Jane wanted to call the dog, so she made something. The object was a shell that now had its tip broken. John was looking for something to play with and inadvertently grabbed the shell. He put its tip in his mouth and blew. 

Whistle object, shortened compromised action scenario: 

Jane wanted to call the dog, so she made something. The object was a shell that now had its tip broken. John wanted to call the dog, so he picked up the object. He put the object to his mouth and whispered the dog’s name. 

\subsection{Experiment E: Percent yes, Yes/No, with in-context learning}

This is the version of the experiment reported in the main paper. This experiment is much like experiment B, except for a few differences. First, the introduction is much more substantial, consisting of approximately 900 words instead of approximately 40. This consists of a description of the experiment, three example warm-up questions about a different domain, and four examples (one positive and one negative for each of two objects – a hanger and a gardening fork) that demonstrate responses in the relevant domain. This material was all drawn from the original text presented to the human participants by Chaigneau et al. After the introduction the uncompromised scenario is presented, the response is recorded and accumulated in the prompt, and the compromised prompts are then presented in random order with these scenarios and their responses also accumulating in successive prompts. Again we record the probability that the model assigns to ``yes'' or versions thereof as the output. In order to control for possible effects from the particular random order in which the compromised prompts are presented, the results are averaged over three repetitions. The hypothesis was that the full introduction text may aid the model in competently completing the task, and that the accumulation of previous responses may aid the model in comparing salient differences between the compromised scenarios and other versions of it. 

Below is an example prompt that would be served to GPT-3 as the fifth prompt in the mop object naming question series. First the introduction and baseline scenario are served to the model for a response, after which prior scenarios and its responses to them accumulate in successive prompts. Note that the responses are standardized for inclusion in subsequent versions of the prompt, so if the model responds with ``YES'' and provides a probability greater than 50\% to versions of ``yes'', then the standard response ``yes'' will accumulate in the prompt. Also note that the warm-up questions in the introduction are altered to be relevant to the live questions in the experiment. The warm-up questions included below for the hanger and gardening fork are for the naming question. For the function question, these would change to ``Does this scenario illustrate the function of a [gardening fork/hanger]''. Likewise for the causality question, these would change to ``Is it likely that, as a result of the events described above, John [loosened the garden soil/hanged his T shirt]''.

Results from GPT-3 on this experiment are shown in Figure \ref{yesnoic_heatmaps_paper}, column 6.

``In this study we are studying knowledge of daily life activities. On the next pages you will find several short stories or scenarios, describing events in the daily lives of two characters. Each scenario is different from the others.

Read each one carefully. As you read them, imagine to yourself what is happening. In each scenario, there are two characters. The names of the two characters in all the scenarios are Jane and John. Jane has a young child, called John, who is constantly running around the house and trying new things.

Here are three examples that will help you to get used to this procedure.

First example:

Jane was reading a book that she was very interested in, and remembered she had to make a phone call. While she made the phone call, she left her book on the table. While Jane was on the phone, John looked for something to do. He saw the book that Jane was reading, picked it up, and browsed through it.

Question:

Does this scenario illustrate a situation where Jane is going to be mad at John?

Answer (respond with yes or no): no

Here is a second scenario:

On a certain occasion Jane was feeling stressed, extremely tired, and grouchy. To get her mind off her worries, she decided to watch some TV. While Jane was watching TV, John started to complain that he was hungry. Reluctantly, Jane had to get up and fix him something to eat.

Question:

Does this scenario illustrate a situation where Jane is going to be mad at John?

Answer (respond with yes or no): no

Here is a third scenario:

One day, Jane was getting ready to go out. She had decided to wear a nice outfit that she had recently bought. She left the outfit over her bed, and took a shower. While Jane was taking a shower, John came into the room carrying a permanent marker he found elsewhere. He decided to climb onto the bed, and in doing so he stained Jane’s outfit.

Question:

Does this scenario illustrate a situation where Jane is going to be mad at John?

Answer (respond with yes or no): yes

The three examples you just read are about emotions. In contrast, the scenarios that you will read and rate next are not about emotions. However, they have the same structure.

One day Jane wanted to loosen the soil in her garden pots, but she didn’t have anything to do it with. So she decided to make something. She gathered all the materials and made it. When she finished, she left it in the garden so she could use it later. The object had three prongs and a handle. Later that day, John was looking for something to loosen the soil in the garden. He saw the object that Jane had made and thought that it would be good for loosening the soil in the garden. He grabbed the object by the handle and repeatedly pushed the prongs into the garden soil.

Question:

Is it appropriate to call this object a gardening fork?

Answer (respond with yes or no): yes

One day Jane was working in her metal shop, and she decided to make something. She gathered all the materials and started to work. As she worked, metal scraps were left to the side of her table. One of the scrap pieces looked like a spherical piece of metal with a handle attached to it. Later that day, John was looking for something play with. He saw the object that Jane had made and thought that it would be good for playing with. He grabbed the object from the spherical part, and repeatedly waved it above his head.

Question:

Is it appropriate to call this object a gardening fork?

Answer (respond with yes or no): no

One day Jane was working in her metal shop, and she decided to make something. She gathered all the materials and started to work. As she worked, metal scraps were left to the side of her table. One of the scrap pieces looked like a wire twisted in different ways. Later that day, John was looking for something play with. He saw the object that Jane had made and thought that it would be good for playing with. He grabbed the object from both sides, and repeatedly squeezed it with his hands.

Question:

Is it appropriate to call this object a hanger?

Answer (respond with yes or no): no

One day Jane wanted to hang her clothes, but she didn’t have anything to do it with. So she decided to make something. She gathered all the materials and made it. When she finished, she left it in her room so she could use it later. The object was a long wire shaped like the outline of a person's shoulders, and with a hook on the top. Later that day, John was looking for something to hang his clothes on. He saw the object that Jane had made and thought that it would be good for hanging his clothes on. He grabbed the object and fit it inside his T shirt so that the hook came out through the neck.

Question:

Is it appropriate to call this object a hanger?

Answer (respond with yes or no): yes

One day Jane wanted to wipe up a water spill on the kitchen floor, but she didn’t have anything to do it with. So she decided to make something. She looked around the house for things that would allow her to make an object for wiping up a water spill on the kitchen floor. She gathered all the materials and made it. When she finished, she left it in the kitchen so she could use it later. The object consisted of a bundle of thick cloth attached to a 4-foot long stick. Later that day, John was looking for something to wipe up a water spill on the kitchen floor. He saw the object that Jane had made and thought that it would be good for wiping up a water spill on the kitchen floor. He grabbed the object with the bundle of thick cloth pointing downward and pressed it against the water spill.

Question:

Is it appropriate to call this object a mop?

Answer (respond with yes or no): yes

One day Jane was cleaning the attic. She picked up a bunch of useless things and put them all inside a big cardboard box. Because the box was overflowing, she used a long stick to shove things down. As she did this, something became attached to the stick. Then, Jane carried the box downstairs. She didn’t notice that as she did this, the stick and the thing that was attached to it fell together, as a single object, to the floor. The object consisted of a bundle of thick cloth attached to a 4-foot long stick. Later that day, John was looking for something to wipe up a water spill on the kitchen floor. He saw the object that Jane had made and thought that it would be good for wiping up a water spill on the kitchen floor. He grabbed the object with the bundle of thick cloth pointing downward and pressed it against the water spill.

Question:

Is it appropriate to call this object a mop?

Answer (respond with yes or no): yes

One day Jane wanted to wipe up a water spill on the kitchen floor, but she didn’t have anything to do it with. So she decided to make something. She looked around the house for things that would allow her to make an object for wiping up a water spill on the kitchen floor. She gathered all the materials and made it. When she finished, she left it in the kitchen so she could use it later. The object consisted of a bundle of plastic bags attached to a 4-foot long stick. Later that day, John was looking for something to wipe up a water spill on the kitchen floor. He saw the object that Jane had made and thought that it would be good for wiping up a water spill on the kitchen floor. He grabbed the object with the bundle of plastic bags pointing downward and pressed it against the water spill.

Question:

Is it appropriate to call this object a mop?

Answer (respond with yes or no): no

One day Jane wanted to wipe up a water spill on the kitchen floor, but she didn’t have anything to do it with. So she decided to make something. She looked around the house for things that would allow her to make an object for wiping up a water spill on the kitchen floor. She gathered all the materials and made it. When she finished, she left it in the kitchen so she could use it later. The object consisted of a bundle of thick cloth attached to a 4-foot long stick. Later that day, John was looking for something to wipe up a water spill on the kitchen floor. He saw the object that Jane had made and thought that it would be good for wiping up a water spill on the kitchen floor. He grabbed the object with the bundle of thick cloth pointing upward instead of downward, and pressed the bare wood end against the water spill.

Question:

Is it appropriate to call this object a mop?

Answer (respond with yes or no): no

One day Jane wanted to wipe up a water spill on the kitchen floor, but she didn’t have anything to do it with. So she decided to make something. She looked around the house for things that would allow her to make an object for wiping up a water spill on the kitchen floor. She gathered all the materials and made it. When she finished, she left it in the kitchen so she could use it later. The object consisted of a bundle of thick cloth attached to a 4-foot long stick. Later that day, John was in the kitchen looking for something to eat. He was distracted as he looked for something and inadvertently grabbed the object that Jane had left in the kitchen. He grabbed the object with the bundle of thick cloth pointing downward and pressed it against the water spill.

Question:

Is it appropriate to call this object a mop?

Answer (respond with yes or no):''

\subsection{Experiment F: Ranked choice}

In this experiment, we prompt the LLM to rank the scenario versions. First an introduction is given, such as (paragraph breaks indicated by ‘/’): “Instruction: choose the statement in which it is most appropriate to call the object a mop./Your choice (give only the number):”. Then, the five prompts are presented in random numbered order. Last, output is requested: “Your choice (give only the number):”. The numerical response is recorded, the relevant option is removed from the list, and the remaining numbers are adjusted accordingly. This is repeated until a full ranking is obtained. In order to control for possible effects from the particular random ordering of the scenario versions, the average rank of each scenario is computed across ten repetitions. 

Below are two illustrative prompts that the model could be served for the mop object and causality question. As can be seen, the third option is removed after it is selected by the model (the response ``3'' is not included in the first prompt to the model, but is shown here for clarity). This process is repeated until the model is presented with a choice between the final two scenarios. The original order of the scenarios is randomized. The first sentence of the prompt is changed for the naming and function questions, in which case it would read as either ``Instruction: choose the statement in which it is most appropriate to call the object a mop'' or ``Instruction: choose the statement that best illustrates the function of a mop''. 

``Instruction: choose the statement according to which it is likely that, as a result of the events described, John wiped up the water spill.
1. One day Jane wanted to wipe up a water spill on the kitchen floor, but she didn’t have anything to do it with. So she decided to make something. She looked around the house for things that would allow her to make an object for wiping up a water spill on the kitchen floor. She gathered all the materials and made it. When she finished, she left it in the kitchen so she could use it later. The object consisted of a bundle of plastic bags attached to a 4-foot long stick. Later that day, John was looking for something to wipe up a water spill on the kitchen floor. He saw the object that Jane had made and thought that it would be good for wiping up a water spill on the kitchen floor. He grabbed the object with the bundle of plastic bags pointing downward and pressed it against the water spill.
2. One day Jane wanted to wipe up a water spill on the kitchen floor, but she didn’t have anything to do it with. So she decided to make something. She looked around the house for things that would allow her to make an object for wiping up a water spill on the kitchen floor. She gathered all the materials and made it. When she finished, she left it in the kitchen so she could use it later. The object consisted of a bundle of thick cloth attached to a 4-foot long stick. Later that day, John was looking for something to wipe up a water spill on the kitchen floor. He saw the object that Jane had made and thought that it would be good for wiping up a water spill on the kitchen floor. He grabbed the object with the bundle of thick cloth pointing downward and pressed it against the water spill.
3. One day Jane was cleaning the attic. She picked up a bunch of useless things and put them all inside a big cardboard box. Because the box was overflowing, she used a long stick to shove things down. As she did this, something became attached to the stick. Then, Jane carried the box downstairs. She didn’t notice that as she did this, the stick and the thing that was attached to it fell together, as a single object, to the floor. The object consisted of a bundle of thick cloth attached to a 4-foot long stick. Later that day, John was looking for something to wipe up a water spill on the kitchen floor. He saw the object that Jane had made and thought that it would be good for wiping up a water spill on the kitchen floor. He grabbed the object with the bundle of thick cloth pointing downward and pressed it against the water spill.
4. One day Jane wanted to wipe up a water spill on the kitchen floor, but she didn’t have anything to do it with. So she decided to make something. She looked around the house for things that would allow her to make an object for wiping up a water spill on the kitchen floor. She gathered all the materials and made it. When she finished, she left it in the kitchen so she could use it later. The object consisted of a bundle of thick cloth attached to a 4-foot long stick. Later that day, John was looking for something to wipe up a water spill on the kitchen floor. He saw the object that Jane had made and thought that it would be good for wiping up a water spill on the kitchen floor. He grabbed the object with the bundle of thick cloth pointing upward instead of downward, and pressed the bare wood end against the water spill.
5. One day Jane wanted to wipe up a water spill on the kitchen floor, but she didn’t have anything to do it with. So she decided to make something. She looked around the house for things that would allow her to make an object for wiping up a water spill on the kitchen floor. She gathered all the materials and made it. When she finished, she left it in the kitchen so she could use it later. The object consisted of a bundle of thick cloth attached to a 4-foot long stick. Later that day, John was in the kitchen looking for something to eat. He was distracted as he looked for something and inadvertently grabbed the object that Jane had left in the kitchen. He grabbed the object with the bundle of thick cloth pointing downward and pressed it against the water spill.
Your choice (give only the number):3''

``Instruction: choose the statement according to which it is likely that, as a result of the events described, John wiped up the water spill.
1. One day Jane wanted to wipe up a water spill on the kitchen floor, but she didn’t have anything to do it with. So she decided to make something. She looked around the house for things that would allow her to make an object for wiping up a water spill on the kitchen floor. She gathered all the materials and made it. When she finished, she left it in the kitchen so she could use it later. The object consisted of a bundle of plastic bags attached to a 4-foot long stick. Later that day, John was looking for something to wipe up a water spill on the kitchen floor. He saw the object that Jane had made and thought that it would be good for wiping up a water spill on the kitchen floor. He grabbed the object with the bundle of plastic bags pointing downward and pressed it against the water spill.   
2. One day Jane wanted to wipe up a water spill on the kitchen floor, but she didn’t have anything to do it with. So she decided to make something. She looked around the house for things that would allow her to make an object for wiping up a water spill on the kitchen floor. She gathered all the materials and made it. When she finished, she left it in the kitchen so she could use it later. The object consisted of a bundle of thick cloth attached to a 4-foot long stick. Later that day, John was looking for something to wipe up a water spill on the kitchen floor. He saw the object that Jane had made and thought that it would be good for wiping up a water spill on the kitchen floor. He grabbed the object with the bundle of thick cloth pointing downward and pressed it against the water spill.     
3. One day Jane wanted to wipe up a water spill on the kitchen floor, but she didn’t have anything to do it with. So she decided to make something. She looked around the house for things that would allow her to make an object for wiping up a water spill on the kitchen floor. She gathered all the materials and made it. When she finished, she left it in the kitchen so she could use it later. The object consisted of a bundle of thick cloth attached to a 4-foot long stick. Later that day, John was looking for something to wipe up a water spill on the kitchen floor. He saw the object that Jane had made and thought that it would be good for wiping up a water spill on the kitchen floor. He grabbed the object with the bundle of thick cloth pointing upward instead of downward, and pressed the bare wood end against the water spill.
4. One day Jane wanted to wipe up a water spill on the kitchen floor, but she didn’t have anything to do it with. So she decided to make something. She looked around the house for things that would allow her to make an object for wiping up a water spill on the kitchen floor. She gathered all the materials and made it. When she finished, she left it in the kitchen so she could use it later. The object consisted of a bundle of thick cloth attached to a 4-foot long stick. Later that day, John was in the kitchen looking for something to eat. He was distracted as he looked for something and inadvertently grabbed the object that Jane had left in the kitchen. He grabbed the object with the bundle of thick cloth pointing downward and pressed it against the water spill.
Your choice (give only the number):''

\subsection{Experiment G: Percent yes, Yes/No, with in-context learning, no logprobs}

In Experiment E we directly record the probability assigned by GPT-3 to ``yes'' or versions thereof. Following the release of GPT-4, we wish to run a similar experiment on the new model but lack direct access to the probability assigned to a generated token. 

Therefore, Experiment G is the same as experiment E except we replace the probability assigned to ``yes'' or versions thereof with the proportion across repeated responses in which ``yes'' or versions thereof are generated as the output. The experiment is repeated for two runs with 5 responses per question averaged in each run. Splitting the ten responses per question across the two runs controls for the particular random order in which the questions are presented in a given run. 

Results from GPT-4 on this experiment are shown in Figure \ref{yesnoic_heatmaps_paper}, column 4.

\subsection{Experiment H: Likert scale with in-context learning}

We observe a negative result using a Likert scale and GPT-3 in experiment A, but following the release of GPT-4 we re-open such an approach. We combine the Likert scale of Experiment A with the in-context learning of experiment E to produce this experiment, which is the most faithful to \citet{Chaigneau:04}. In particular, we take exactly the prompts from Experiment E, except we replace all references to Yes/No responses with references to Likert scale responses. So we replace a sentence such as ``Answer (respond with yes or no): no'' with a sentence such as ``Answer (respond with only a number between 1 and 7 inclusive, with 1 being ``not at all'' and 7 being ``very well''): 1''. We replace the responses to the warm-up introduction questions, which were (no, no, yes, yes, no, no, yes), with numeric responses (1, 2, 7, 6, 1, 2, 7). These numeric responses were arrived at by applying our own judgment to the questions (in the original experiment, human subjects discussed these with each other in order to gain familiarity with using the scale). Similarly, when prompting the model for a response, we replace the final ``Answer (respond with yes or no):'' with ``Answer (respond with only a number between 1 and 7 inclusive, with 1 being ``not at all'' and 7 being ``very well''):''. As in Experiment E, we accumulate the model's own responses to prior questions in subsequent prompts. 

We obtain repeated outputs and average the Likert responses across them. The experiment is repeated for two runs with 5 responses per question averaged in each run. Splitting the ten responses per question across the two runs controls for the particular random order in which the questions are presented in a given run. 

Results from GPT-4 on this experiment are shown in Figure \ref{yesnoic_heatmaps_paper}, column 3. Results from GPT-3 on this experiment are shown in Figure \ref{yesnoic_heatmaps_paper}, column 5.

\subsection{Experiment I: Likert scale with in-context learning, Experiment 2}

This experiment is referred to as Experiment 2 in the paper, and corresponds to Experiment 2 of \citet{Chaigneau:04}. 

The setup is identical to experiment H above, except that in Experiment H all factors are compromised one at a time, whereas in Experiment I we only vary one or both of History and Goal. Further, experiment H is conducted using the Causal, Function, and Naming questions. Following \citet{Chaigneau:04}, we only use the Causal and Function questions for this experiment. 

\section{Additional data: Heatmaps showing data from different experiment versions and correlations between experiments}
\label{sec:additional_data}

\begin{figure}[ht]
\includegraphics[width=\linewidth]{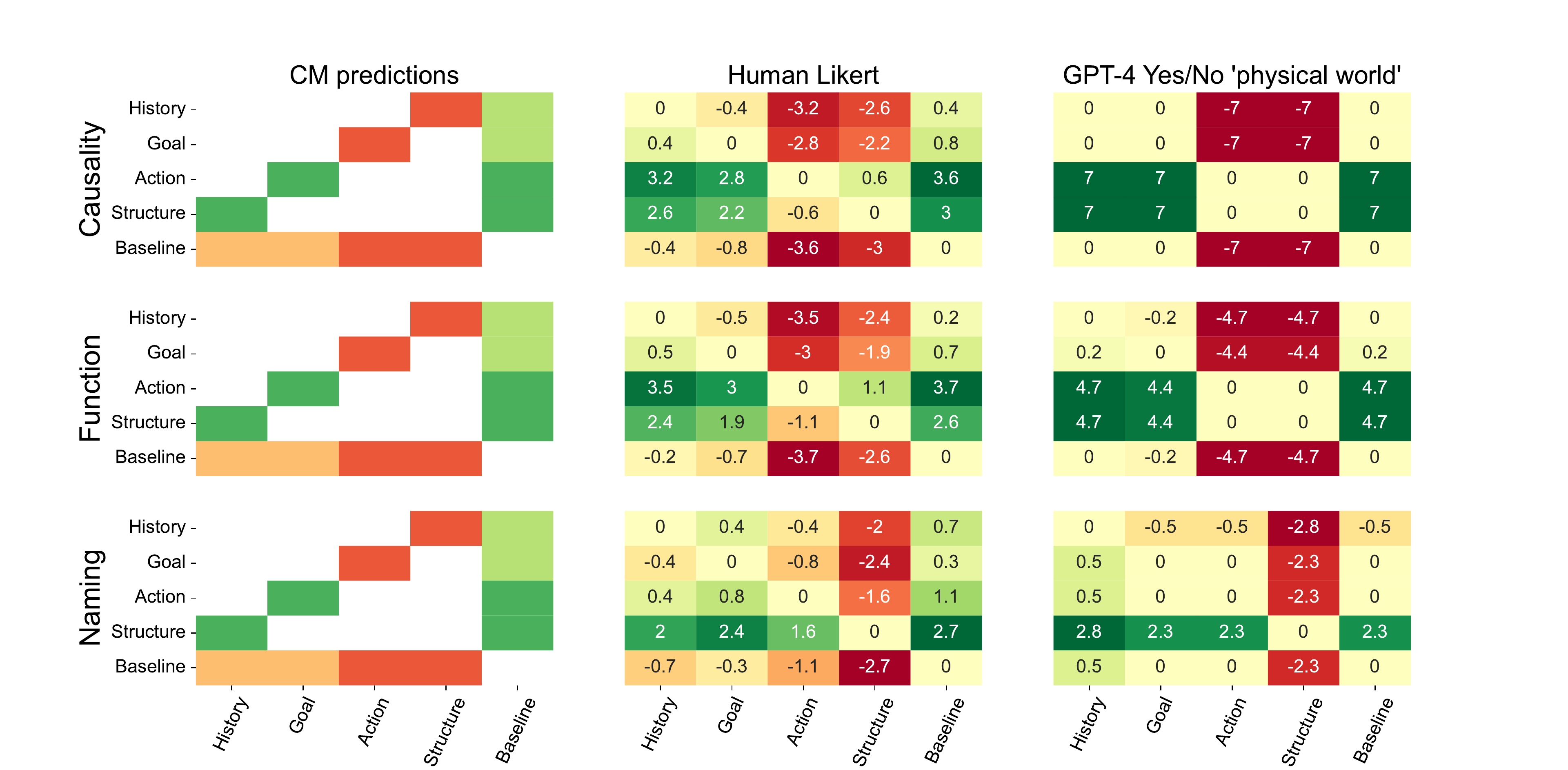}
\caption{\label{gpt4_pw_heatmap}Heatmap showing data from the ``Percent yes, Yes/No, with in-context learning, no logprobs'' version of the experiment, conducted on GPT-4 with the system prompt ``You are a helpful assistant with an excellent understanding of the physical world''.}
\end{figure}

\begin{figure}[ht]
\includegraphics[width=\linewidth]{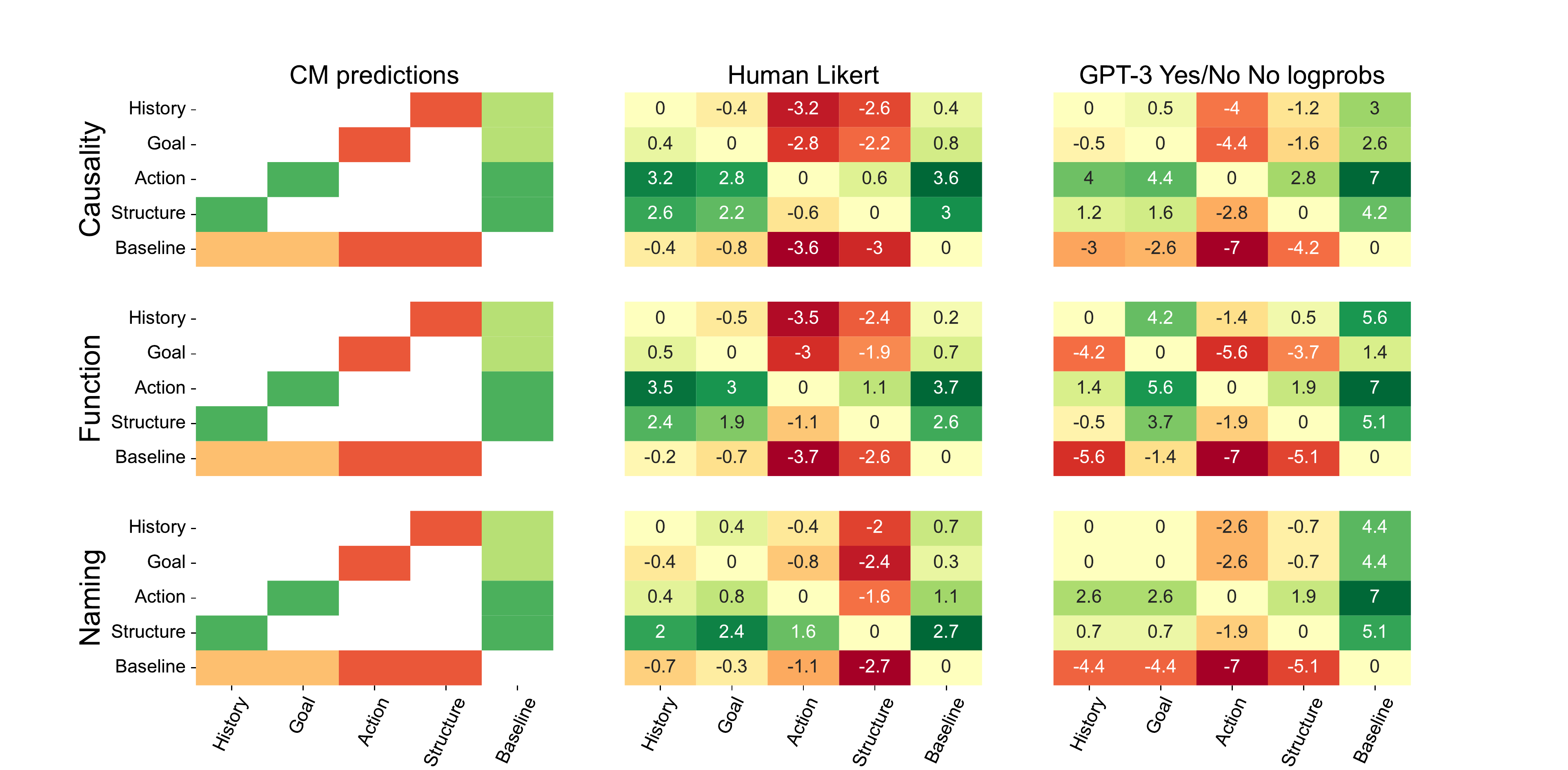}
\caption{\label{gpt3nlp_heatmap}Heatmap showing data from the ``Percent yes, Yes/No, with in-context learning, no logprobs'' version of the experiment (GPT-3).}
\end{figure}

\begin{figure}[ht]
\includegraphics[width=\linewidth]{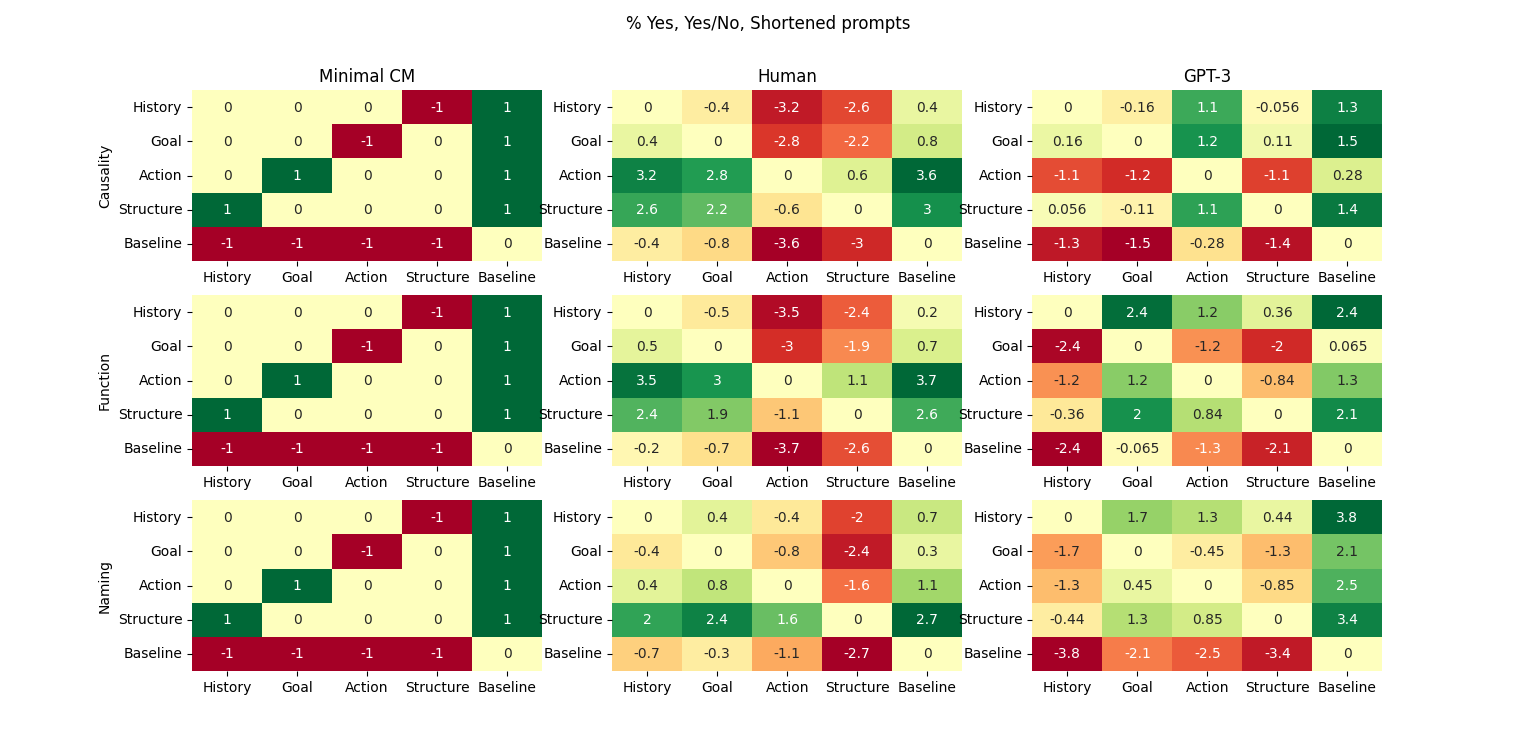}
\caption{\label{yesnoshort_heatmap}Heatmap showing data from the ``Percent yes, Yes/No, shortened prompts'' version of the experiment (GPT-3).}
\end{figure}

\begin{figure}[ht]
\includegraphics[width=\linewidth]{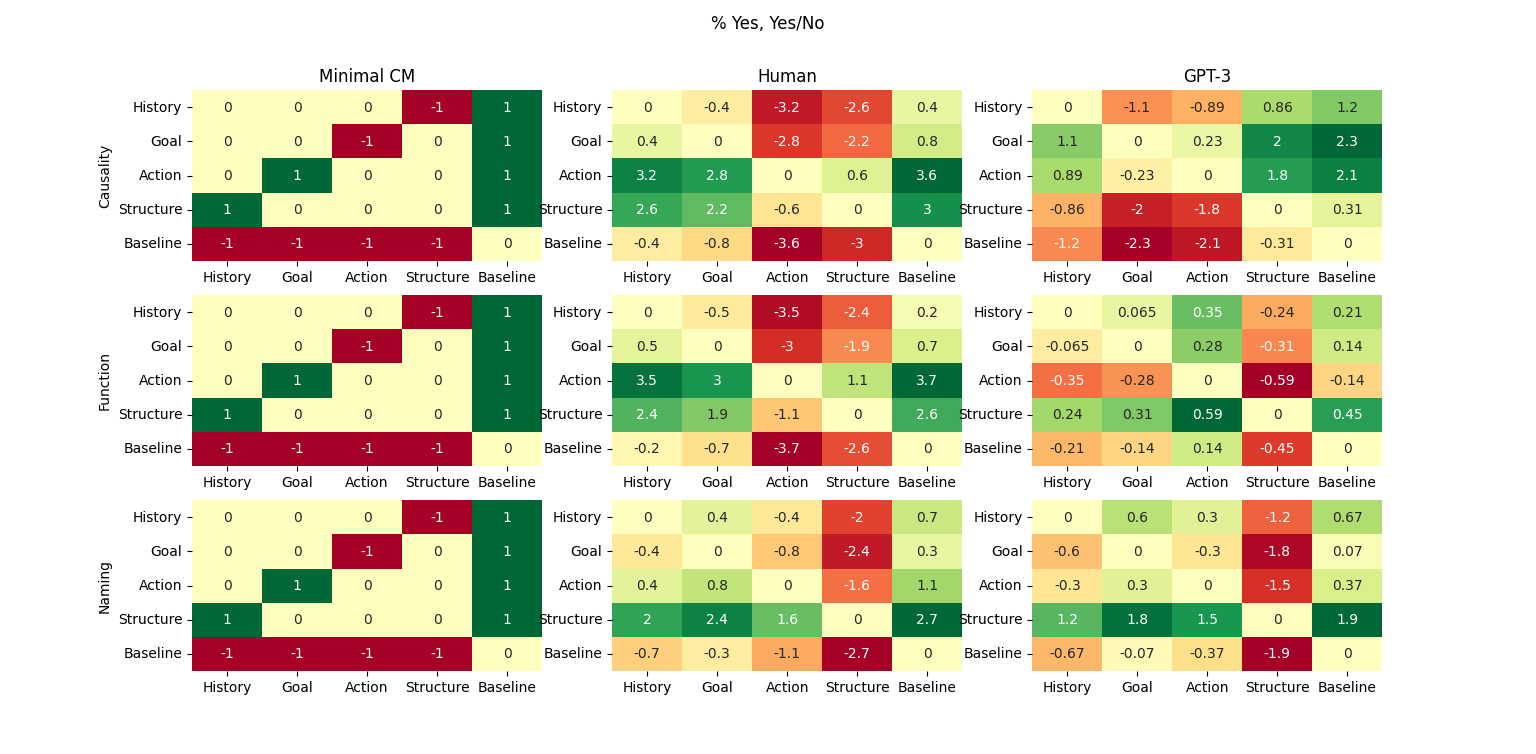}
\caption{\label{yesno_heatmap}Heatmap showing data from the ``Percent yes, Yes/No'' version of the experiment (GPT-3).}
\end{figure}

\begin{figure}[ht]
\includegraphics[width=\linewidth]{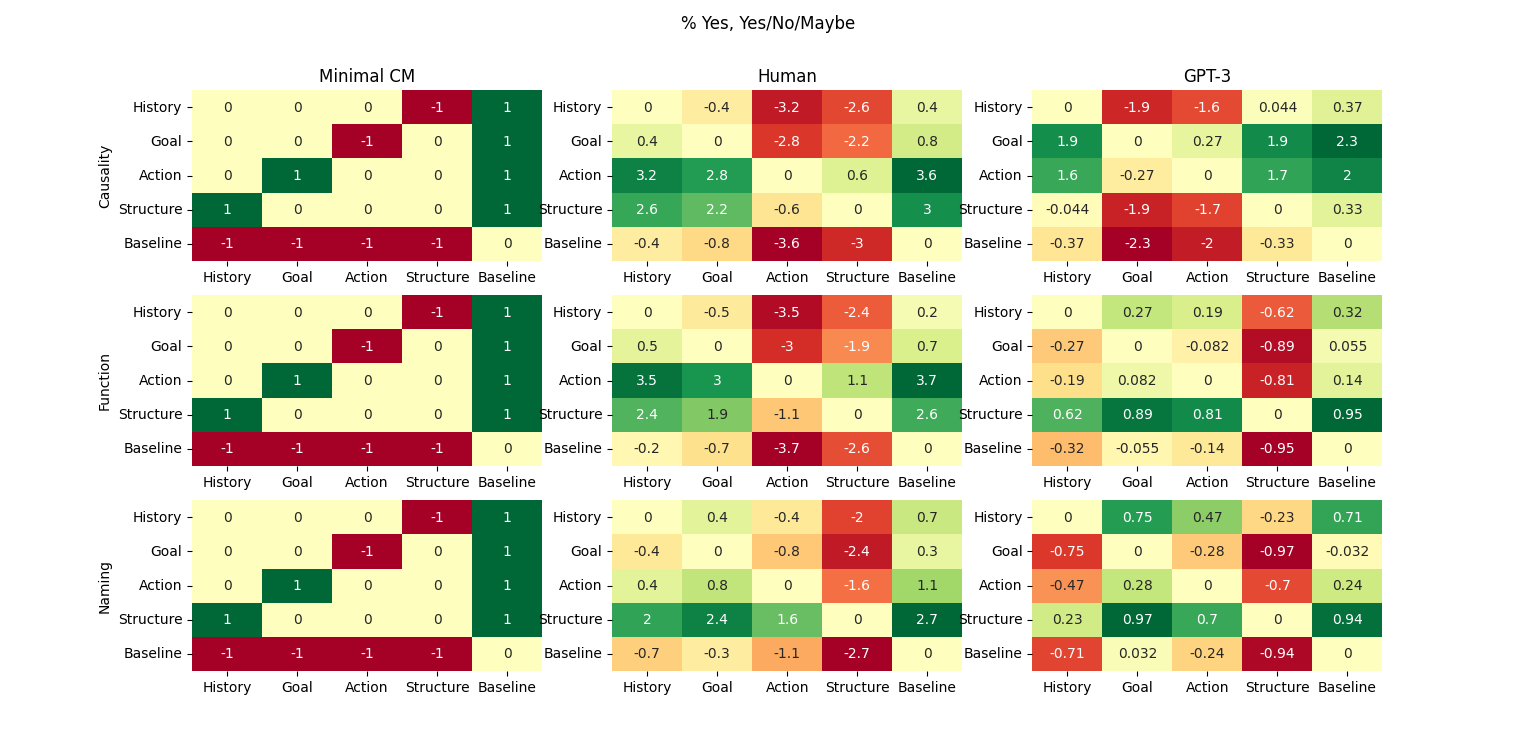}
\caption{\label{yesnomaybe_heatmap}Heatmap showing data from the ``Percent yes, Yes/No/maybe'' version of the experiment (GPT-3).}
\end{figure}

\begin{figure}[ht]
\includegraphics[width=\linewidth]{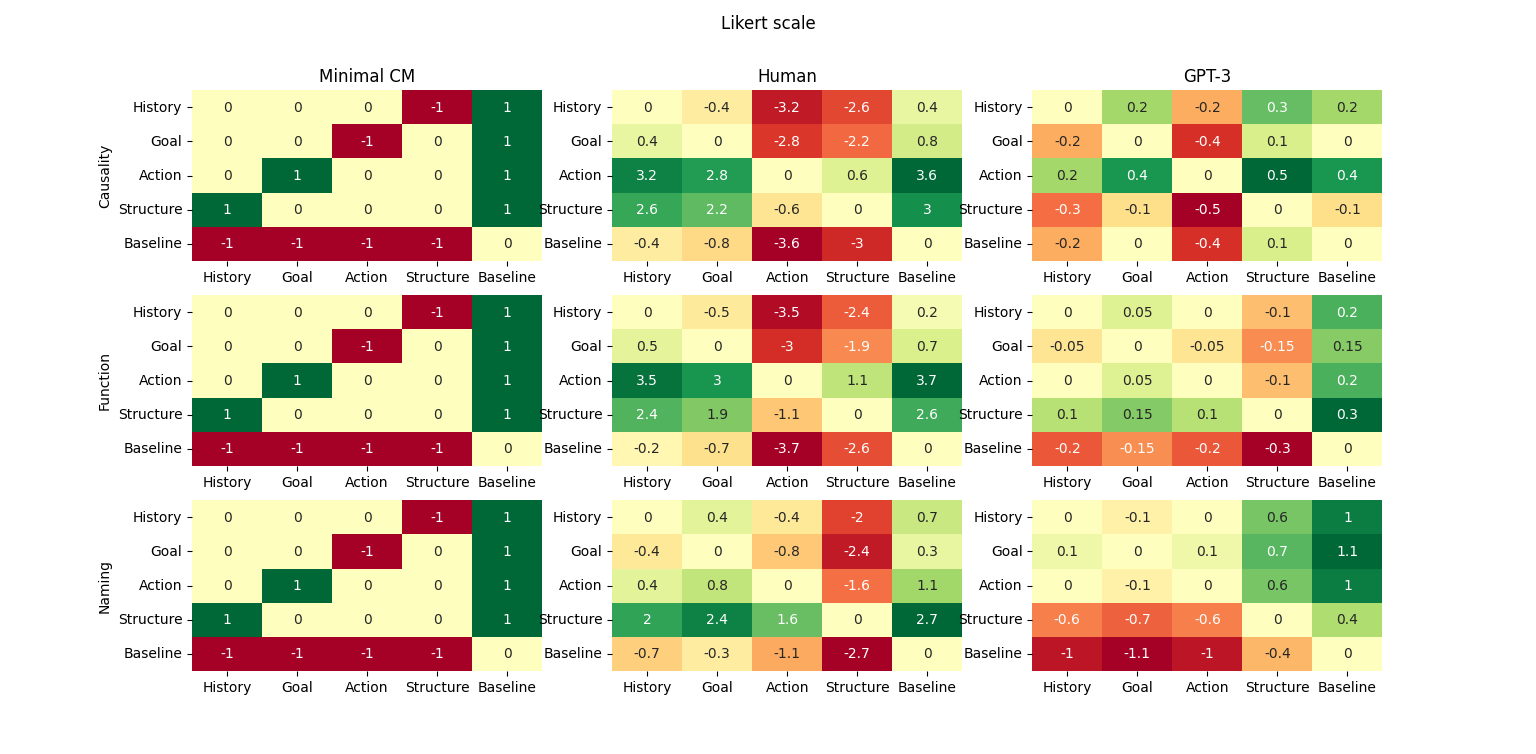}
\caption{\label{likert_heatmap}Heatmap showing data from the ``Likert scale'' version of the experiment. Note that this experiment was implemented for the mop object only (GPT-3).}
\end{figure}

\begin{figure}[ht]
\includegraphics[width=\linewidth]{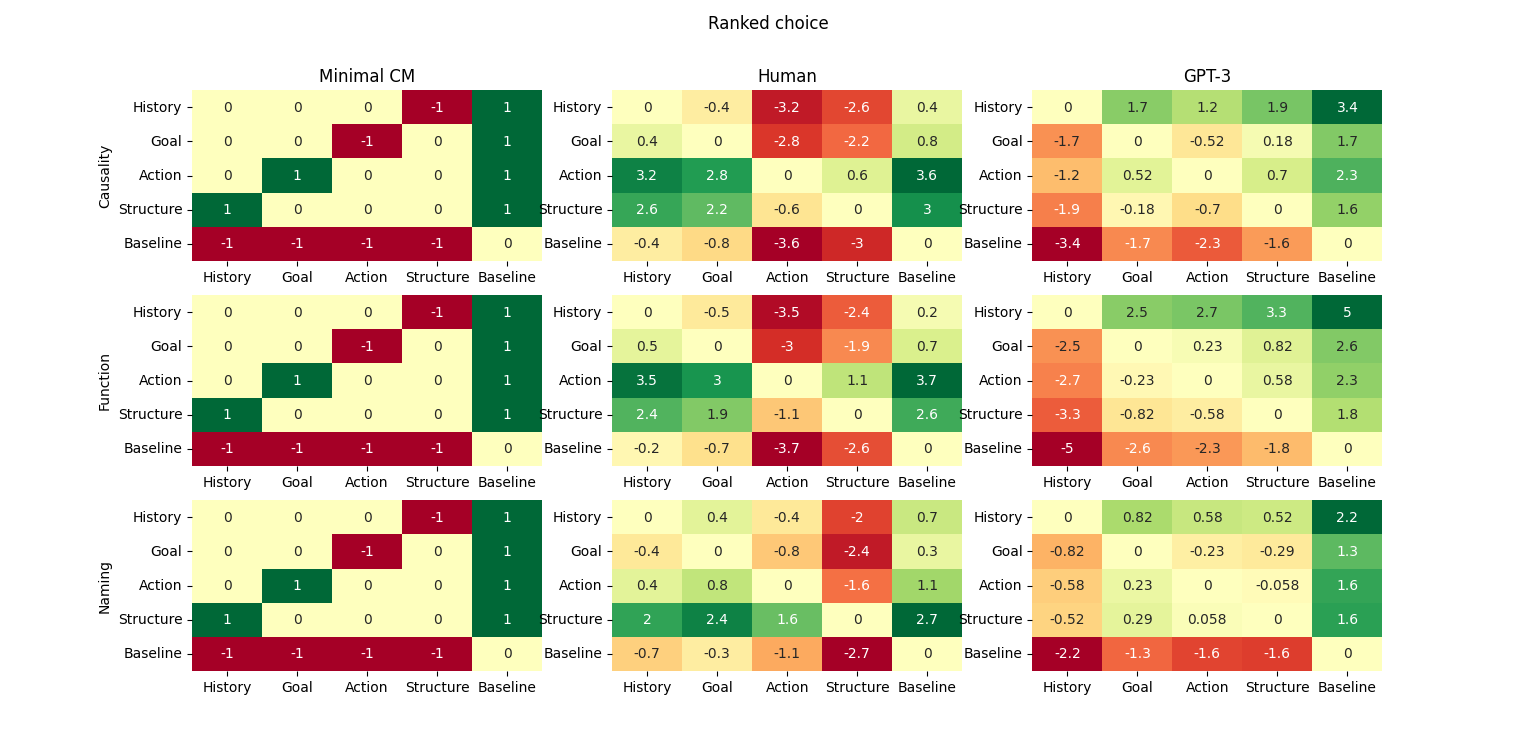}
\caption{\label{rc_heatmap}Heatmap showing data from the ``Ranked choice'' version of the experiment (GPT-3).}
\end{figure}

\begin{figure*}
  \includegraphics[width=\linewidth]{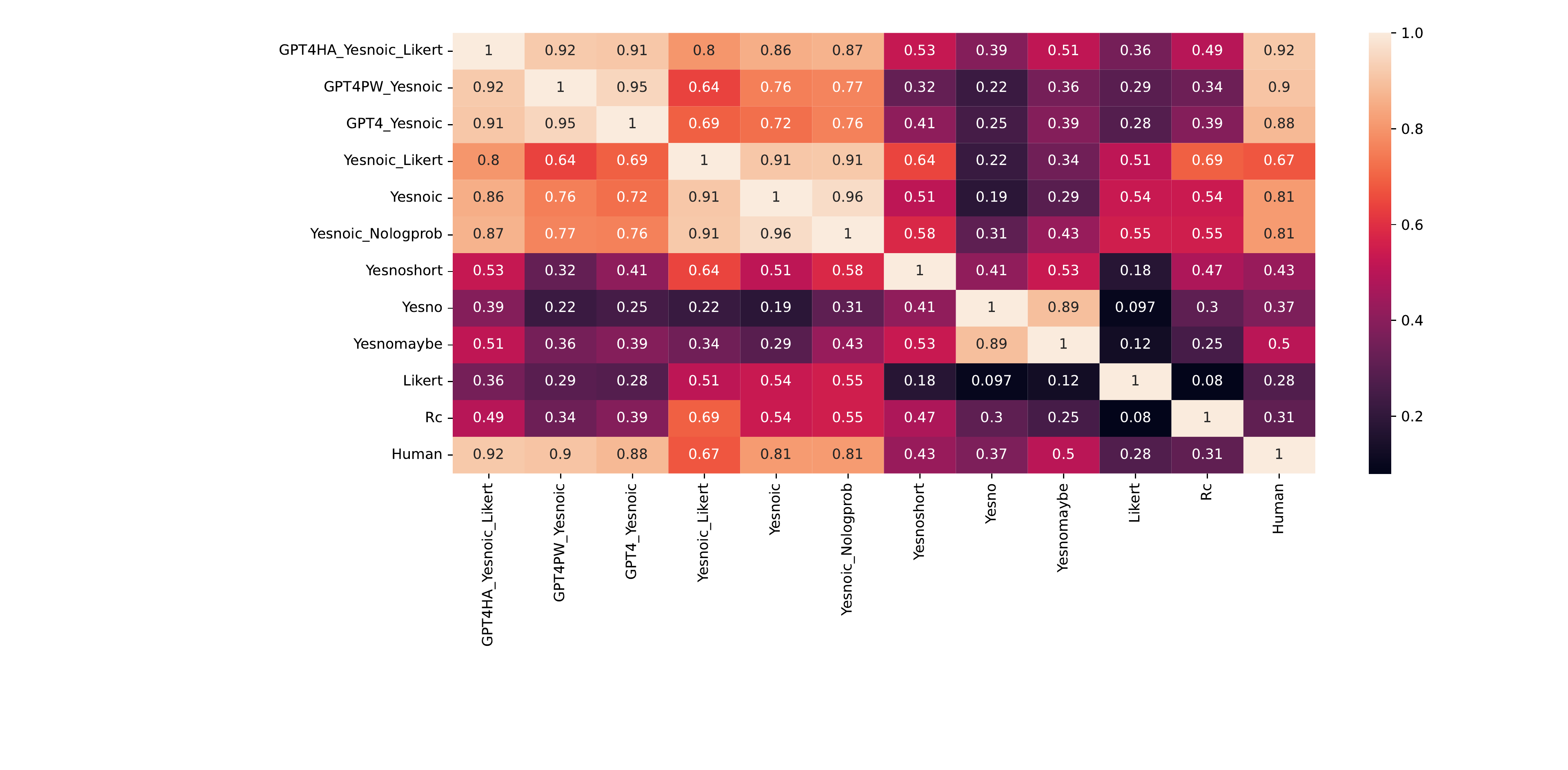}
  \captionof{figure}{\label{experiment_corrs}The Spearman correlation between data from the different versions of the experiment. When not labeled as either `Human' or `GPT-4', the experiment is conducted on GPT-3. GPT4PW indicates that the system prompt for GPT-4 includes that the model has ``an excellent understanding of the physical world''. The multiple experiment versions conducted on GPT-3 strengthen our negative results for that model, as we are not able to find a positive result despite providing it with numerous opportunities under different response modalities and prompting methods to respond to the experiment as predicted by HIPE. The experiments shown here correspond in order to Experiments H, G, G, H, E, G, D, B, C, A, F, and finally the human data from \citet{Chaigneau:04}}
  \label{exp_corrs}
\end{figure*}

\end{document}